\documentclass{article}

\usepackage{acronym}
\usepackage{xspace}
\usepackage{graphicx}
\usepackage{caption}
\usepackage{amsmath}
\usepackage{array}
\usepackage{xcolor}
\usepackage{subcaption}
\usepackage{booktabs}
\usepackage{comment}
\usepackage{tabularx}
\usepackage{multirow}
\usepackage{float}
\usepackage{dsfont}
\usepackage{siunitx}

\usepackage{enumitem}
\usepackage{wrapfig}
\definecolor{LightGray}{gray}{0.9}
\definecolor{cvprblue}{rgb}{0.21,0.49,0.74}
\usepackage[pagebackref,breaklinks,colorlinks=true,linkcolor=cvprblue,citecolor=cvprblue,bookmarks=true]{hyperref}
\usepackage[capitalise,nameinlink]{cleveref}
\usepackage[sort]{cite}
\usepackage{multicol}

\makeatletter
\renewcommand{\paragraph}{%
  \@startsection{paragraph}{4}%
  {\z@}{0ex \@plus 0ex \@minus 0ex}{-1em}%
  {\hskip\parindent\normalfont\normalsize\bfseries}%
}
\makeatother

\newcommand{\eg}{\emph{e.g.}\xspace}

\acrodef{ddim}[DDIM]{Denoising Diffusion Implicit Models}
\acrodef{mlp}[MLP]{multi-layer perception}
\acrodef{bc}[BC]{Behavior cloning}
\acrodef{mpc}[MPC]{model predictive control}
\acrodef{rl}[RL]{reinforcement learning}
\acrodef{il}[IL]{imitation learning}
\acrodef{wbc}[WBC]{Whole body control}
\acrodef{llm}[LLM]{large language model}
\acrodef{mmlm}[MMLM]{multi-modal language model}
\acrodef{sft}[SFT]{supervised fine-tuning}
\acrodef{rlhf}[RLHF]{Reinforcement Learning from Human Feedback}
\acrodef{sg3d}[3DSG]{3D Scene Graph}
\acrodef{qa}[QA]{question-answering}
\acrodef{gui}[GUI]{graphical user interface}
\acrodef{cot}[CoT]{Chain-of-Thought}
\acrodef{mse}[MSE]{mean squared error}
\acrodef{ppo}[PPO]{Proximal Policy Optimization}

\newcommand{\numtask}{7\xspace}

\definecolor{gblue}{HTML}{4285F4}
\definecolor{gred}{HTML}{DB4437}
\definecolor{ggreen}{HTML}{0F9D58}
\definecolor{vblue}{HTML}{2993ba}

\usepackage{amsmath,amsfonts,bm}

\def\eqref#1{equation~\ref{#1}}
\def\1{\bm{1}}

\def\rvtheta{{\mathbf{\theta}}}
\def\rvomega{{\mathbf{\omega}}}
\def\rva{{\mathbf{a}}}

\def\rvc{{\mathbf{c}}}

\def\rvg{{\mathbf{g}}}

\def\rvo{{\mathbf{o}}}

\def\rvq{{\mathbf{q}}}

\def\rvv{{\mathbf{v}}}

\def\rvx{{\mathbf{x}}}

\def\mF{{\bm{F}}}

\DeclareMathAlphabet{\mathsfit}{\encodingdefault}{\sfdefault}{m}{sl}
\SetMathAlphabet{\mathsfit}{bold}{\encodingdefault}{\sfdefault}{bx}{n}

\usepackage{amssymb}

\usepackage[final]{corl_2025} % Uncomment for the camera-ready ``final'' version.
\title{Learning a Unified Policy for Position and Force Control in Legged Loco-Manipulation}

\author{
Peiyuan Zhi$^{1,2,*}$, Peiyang Li$^{1,3,*}$, Jianqin Yin$^{3}$, Baoxiong Jia$^{1,2,\dagger}$, Siyuan Huang$^{1,2,\dagger}$ \\
$^{1}$ State Key Laboratory of General Artificial Intelligence, BIGAI \\
$^{2}$ Joint Laboratory of Embodied AI and Humanoid Robots, BIGAI \& UniTree Robotics \\
$^{3}$ Beijing University of Posts and Telecommunications \\
\texttt{\url{https://unified-force.github.io/}} \\
{\footnotesize $^{*}$ Equal contribution. \quad $^\dagger$ Corresponding authors.}
}
\begin{document}
\maketitle

\begin{center}
    \centering
    \vspace{-25pt}
    \captionsetup{type=figure}
        % \fbox{\rule[0cm]{0cm}{5cm}\rule[0cm]{\linewidth}{0cm}}
        % \fbox{\rule[0cm]{0cm}{8cm}\rule[0cm]{\linewidth}{0cm}}
        \includegraphics[width=\linewidth]{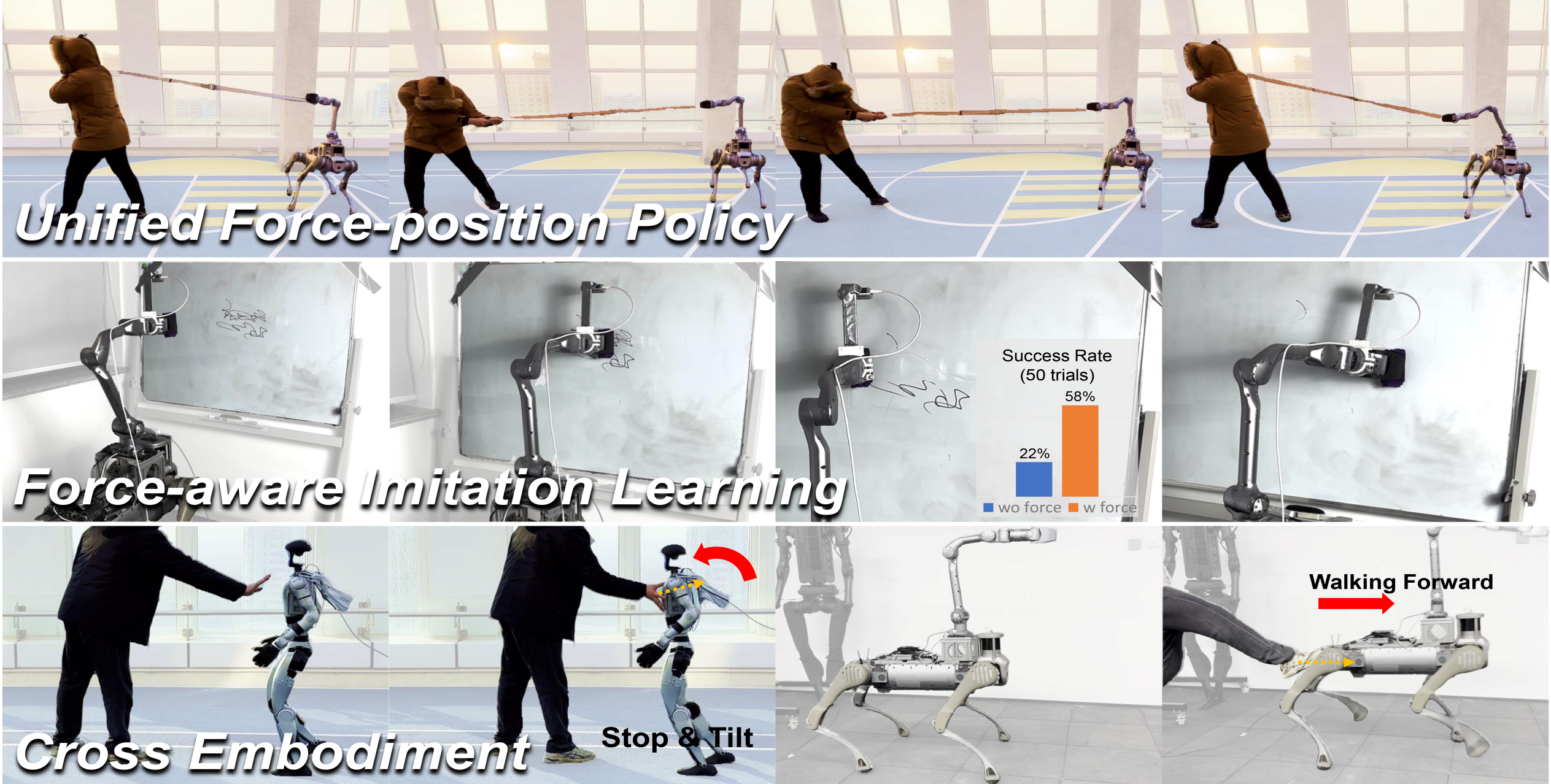}
        \captionof{figure}{
        \textbf{We present a unified force-position policy for legged robots} that enables diverse loco-manipulation behaviors, including position tracking, force application, and compliant interactions (top). When used for imitation learning data collection, the policy's learned internal force estimator provides force-aware demonstrations, improving model performance in contact-rich tasks without external force sensors (middle). Results on quadruped and humanoid robots demonstrate the policy's versatility and robustness (bottom).
        %An overview of the unified force-position policy for legged robots} that enables simultaneous force and position control. This policy supports a wide range of loco-manipulation behaviors, including position tracking, force application, and compliant interaction (top). Additionally, it can be integrated into imitation learning pipelines by leveraging force estimation to enhance contact-rich manipulation tasks (middle). The effectiveness of the policy is validated on both quadrupedal and humanoid robots, demonstrating its versatility and robustness across different morphologies and task settings (bottom).
        }
    \label{fig:teaser}
\end{center}

%===============================================================================

\vspace{-0.20in}

\begin{abstract}
Robotic loco-manipulation often involves contact-rich interactions with the environment, requiring the joint modeling of contact force and robot position.
However, recent visuomotor policies often focus solely on learning position or force control, overlooking their co-learning. We propose the first unified policy for legged robots that jointly models force and position control learned without relying on force sensors. By simulating diverse combinations of position and force commands alongside external disturbance forces, we use \acl{rl} to learn a policy that estimates forces from historical robot states and compensates for them through position and velocity adjustments. This policy enables a wide range of manipulation behaviors under varying force and position inputs, including position tracking, force application, force tracking, and compliant interactions. Moreover, we demonstrate that the learned policy enhances trajectory-based imitation learning pipelines by incorporating essential contact information through its force estimation module, achieving approximately $\sim$39.5\% higher success rates in four challenging contact-rich manipulation tasks over position-control policies. Experiments on both a quadrupedal manipulator and a humanoid robot validate the versatility and robustness of the proposed policy in diverse scenarios.
\end{abstract}

\vspace{-0.15in}


%

%===============================================================================

\section{Introduction}\label{sec:intro}

% \todo{half page}

Legged robots have recently advanced in locomotion and manipulation ~\cite{hwangbo2019learning,zhuang2024humanoid,he2024agile,hoeller2024anymal}, enabling them to traverse complex terrains (\eg, stairs) and extend their workspace through adaptive body posture, revitalizing interest in loco-manipulation \cite{sleiman2023versatile,fu2023deep,liu2024visual,qiu2024wildlma}. However, controlling legged manipulators is challenging due to their complex kinematic structures. This difficulty is further exacerbated in contact-rich manipulation tasks, where accurate modeling of contact forces is essential for desired control behaviors (\eg, compliance), yet is hindered by the absence of force-sensing hardware. These challenges underscore the need for robust, adaptable policies to support effective robot-environment and human-robot interactions.

To tackle the control challenge of legged manipulators, \ac{rl} algorithms have emerged as effective alternatives to traditional control methods, offering robust and generalizable policies trained through domain randomization~\cite{fu2023deep,liu2024visual,qiu2024wildlma,portela2024learning,zhuang2024humanoid,zhuang2023robot}. These policies integrate locomotion and manipulation in complex tasks but primarily depend on precise position control, limiting their applicability in contact-rich scenarios. This reliance has also driven the rise of position-based robot imitation learning~\cite{shridhar2023perceiver, chi2023diffusion,brohan2023rt,team2024octo,black2024pi_0}, with large datasets~\cite{walke2023bridgedata, brohan2023rt, khazatsky2024droid, o2024open} focused solely on robot trajectories, omitting crucial contact information due to the lack of force sensing. As shown in~\cref{sec:exp:imitation}, such trajectory-only data is insufficient for training effective policies, even for basic contact-rich tasks (\eg, wiping a blackboard). This underscores the limitation of position control and emphasizes the necessity of integrating force sensing and modeling into learning-based policies for more effective task execution.

In light of the aforementioned challenges and observations, we propose \textbf{the first unified policy for legged robots that seamlessly integrates force and position control without the need for force sensors}. Unlike previous methods~\cite{portela2024learning} that handle force and position control independently, we train a single control policy using \ac{rl} in Isaac Gym~\cite{makoviychuk2021isaac} by simulating diverse combinations of position and force commands alongside external disturbance forces. The policy leverages a force estimator to predict external forces based on the robot's historical states and offsets to target positions, enabling adaptive adjustments to the robot's position and velocity. The learned policy supports versatile manipulation behaviors, including position tracking, force application, force tracking, and compliant responses to varied force and position inputs. We also verify the generalizability of this learning framework to different robot embodiments through an extensive spectrum of \numtask experiments on both the Unitree B2-Z1 quadrupedal manipulator platform and the Unitree G1 humanoid robot.

Additionally, we highlight the capabilities of the learned policy in facilitating imitation learning with contact force information. Specifically, we develop a force-aware data collection pipeline that utilizes our learned policy as the base teleoperation policy, simultaneously passing position and force commands to the robot while collecting contact-rich manipulation data via the embedded contact force estimator. We validate the effectiveness of this data by integrating the estimated force into a position-based imitation learning policy, leading to \textbf{a significant improvement ($\sim$39.5\%) in success rates over the vanilla position-based methods} across three challenging contact-rich tasks. These experimental results underscore the potential of our learned policy as a general framework for curating contact-rich robot interaction data, particularly in the absence of explicit force sensors.

Overall, our contributions can be summarized as follows:

\begin{enumerate}[leftmargin=*,noitemsep,nolistsep]
    \item We propose the first model for learning unified force and position control in legged loco-manipulation, enabling diverse control behaviors such as position tracking, force control, and compliance with a single policy.
    \item Through \numtask experiments on a quadrupedal manipulator and a humanoid robot, we demonstrate the effectiveness and robustness of our learned policy across diverse and challenging task scenarios.
    \item We develop a force-aware robot imitation learning data collection pipeline using our learned force estimator, improving position-based imitation learning baselines by $\sim$39.5\% on three challenging contact-rich manipulation tasks, highlighting our policy's promise as a general and efficient framework for contact-rich task demonstration curation.
\end{enumerate}

%===============================================================================

\section{Related Works}

\paragraph{Whole-body Control}
\ac{wbc} has been widely adopted to enhance robotic capabilities in mobile manipulation, particularly within classical control frameworks \cite{sleiman2021unified,polverini2020multi, sleiman2023versatile}. More recently, \ac{rl} with parallel simulators \cite{makoviychuk2021isaac, rudin2022learning} has become the mainstream approach for addressing complex control challenges in legged robots. Several learning-based methods \cite{fu2023deep,pan2024roboduet, ma2022combining,liu2024visual,wang2024quadwbg} have improved the robustness of \ac{wbc}, while others have extended its application to force-intensive tasks \cite{murphy2012high,murooka2015whole, rehman2016towards,bellicoso2019alma,risiglione2022whole}. For instance, \cite{murooka2015whole} coordinates joint movements to apply sufficient force during pushing, and ALMA\cite{bellicoso2019alma} combines \ac{wbc} with force control to achieve precise end-effector actuation. \cite{risiglione2022whole} integrates Cartesian impedance control into a QP formulation, enabling compliant loco-manipulation through a double mass-damper-spring model. These works collectively highlight \ac{wbc}’s effectiveness in unifying force and position control.

\paragraph{Hybrid Force and Position Control}
In contact-rich manipulation tasks, relying solely on end-effector trajectory control is often insufficient due to the inherent coupling between force and position. Early work \cite{raibert1981hybrid,mason1981compliance,yoshikawa1987dynamic,hogan1984impedance}, including the introduction of impedance control \cite{hogan1984impedance}, laid the foundation for hybrid force-position strategies. Recent studies \cite{hwangbo2019learning,zhang2021learning,hou2024adaptive,de2024current,portela2024learning} have advanced compliance control, with some leveraging force sensors and others estimating force indirectly via internal signals or reinforcement learning. Inspired by these trends, our work eliminates the need for force sensors by using reinforcement learning to train a quadruped robot to simultaneously control force and position. This enables flexible switching between force following, impedance control, and hybrid modes through different command configurations.

\paragraph{Imitation Learning for Mobile Manipulation}
Imitation learning \cite{bousmalis2023robocat,mees2024octo,vuong2023open, yang2023polybot,fang2024rh20t} has recently become a prominent approach for training robots to perform various tasks. \ac{bc} \cite{pomerleau1988alvinn,bojarski2016endendlearningselfdriving} is a straightforward method of imitation learning that learns policies by supervising observation-action pairs from expert demonstrations. Studies leveraging image data and proprioceptive sensing \cite{fu2024mobile,ha2024umi,he2024learning,qiu2024wildlma} to generate robot control commands have shown remarkable success in mobile manipulation tasks. Furthermore, recent research \cite{lin2024learning,yang2023seq2seq} has begun incorporating tactile sensing to enhance the sensory capabilities of robots. Similarly, our work utilizes force inputs without relying on force sensors, demonstrating that force information is critical in enabling robots to complete challenging tasks effectively.

%===============================================================================
\section{Method} \label{sec:method}
\subsection{A Unified Formulation for Force and Position Control}\label{sec:method:formulation}
\begin{figure*}[t!]
    \centering
    \includegraphics[width=\linewidth]{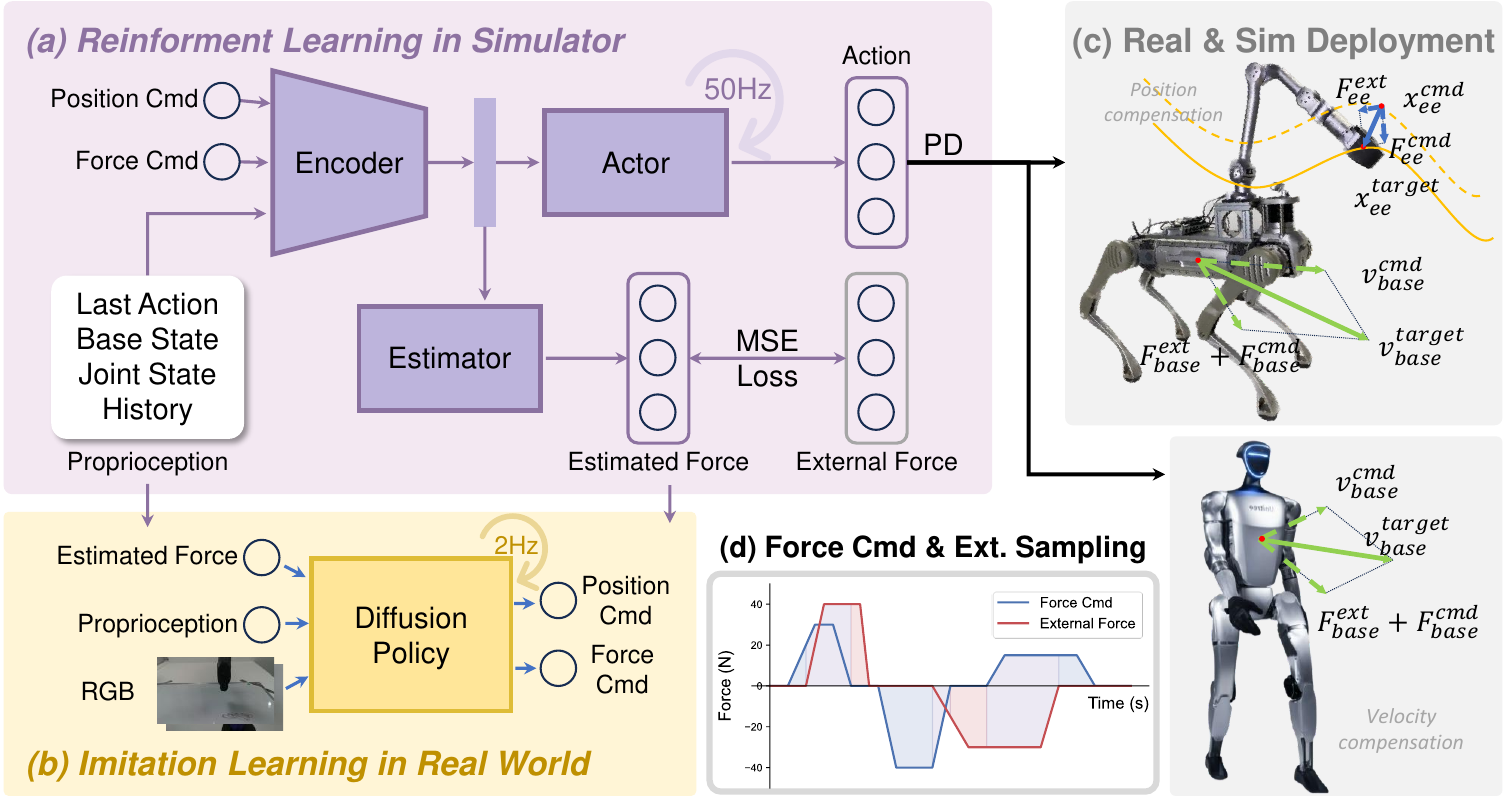}
    \caption{\textbf{Method Overview}. (a) Architecture of the unified position-force policy trained via reinforcement learning to track position and force commands under external disturbances. (b) Force-aware imitation learning enabled by demonstrations collected using our learned policy, without requiring force sensors. (c) Illustration of position and velocity compensation for force interactions modeled at both the end-effector and the robot base. (d) Visualization of sampled force commands and disturbances used to simulate diverse contact scenarios during policy training.
    }
    \label{fig:method:model}
\end{figure*}

We begin by introducing the general problem formulation of our approach. As shown in the upper part in~\cref{fig:method:model}(c), given the position command relative to the robot body frame and force command, $\rvx^{\text{cmd}}$ and $\mF^{\text{cmd}}$, our goal is to learn a \ac{rl} policy that ensures the robot's behavior adheres to these commands under net force $\mF$.
% \footnote{For notational clarity, we omit the subscript $\text{ee}$ from \cref{fig:method:model} in~\cref{sec:method:formulation}.}.
To achieve this goal, we adopt the impedance control formulation:
\begin{equation}
\mF = K(\rvx - \rvx^{\text{des}}) + D(\dot{\rvx} - \dot{\rvx}^{\text{des}}) + M(\ddot{\rvx} - \ddot{\rvx}^{\text{des}}),
\label{eq:impedance_general}
\end{equation}
where $\rvx$ denotes the actual position of the robot. $\rvx^{\text{des}}$, $\dot{\rvx}^{\text{des}}$, and $\ddot{\rvx}^{\text{des}}$ denotes the desired goal position, velocity, and acceleration of the robot. The parameters $K$, $D$, and $M$ correspond to the stiffness and damping coefficients, and equivalent mass (inertia), respectively.

% \subsection{Problem Formulation} \label{sec:method:formulation}
% In this section, we present the general formulation of our approach. As illustrated in~\cref{fig:method:model}(c), we account for forces applied to or exerted by both the end-effector and the robot base. 

% {\textbf{A Unified Formulation for Position and Force Control.} \quad}
% As shown in the upper figure in~\cref{fig:method:model}(c), given the end-effector position command relative to the robot body frame and force command, $\rvx^{\text{cmd}}$ and $\mF^{\text{cmd}}$, our goal is to learn a \ac{rl} policy that ensures the robot's behavior adheres to these commands under net force $\mF$\footnote{For notational clarity, we omit the subscript $\text{ee}$ from \cref{fig:method:model} in~\cref{sec:method:formulation}.}. To achieve this goal, we adopt the impedance control formulation:

\paragraph{End-effector Modeling} As the end-effector typically moves slowly during manipulation tasks, we can make the following simplification over \cref{eq:impedance_general}:
\(\mF = K(\rvx - \rvx^{\text{des}})\).
% \begin{equation}
%     \mF = K(\rvx - \rvx^{\text{des}}).
    % \label{eq:impedance_simplification}
% \end{equation}
The net force $\mF$ primarily consists of three components: the active force $\mF^{\text{cmd}}$, the passive reaction force $\mF^{\text{react}}$ which arises from applying $\mF^{\text{cmd}}$ to the environment, and additional external disturbances $\mF^{\text{ext}}$. Therefore, the desired target position $\rvx^{\text{target}}$ of the end-effector is given by:
\begin{equation}
\rvx^{\text{target}} = \rvx^{\text{cmd}} + \frac{\mF^{\text{ext}} + (\mF^{\text{cmd}} - \mF^{\text{react}})}{K},
\label{eq:pos_force_target}
\end{equation}
where the environment reaction force prevents the end-effector from reaching the commanded position $\rvx^{\text{cmd}}$. Under the formulation of \cref{eq:pos_force_target}, several manipulation behaviors can be derived by appropriately specifying the position command $\rvx^{\text{cmd}}$ and force command $\mF^{\text{cmd}}$, including \textit{Position Control}, \textit{Force Control}, \textit{Impedance Control} and \textit{Hybrid Position and Force Control}. We present a detailed formulation of these control behaviors in~\cref{sec:appendix:formulation}. In complex scenarios involving simultaneous position commands, force commands, and external disturbances, the system adheres to \cref{eq:pos_force_target}, integrating these basic control modes.

% \begin{figure}[t!]
%     \centering
%     \includegraphics[width=0.90\linewidth]{fig/method_target.pdf}
%     \caption{\textbf{Position and velocity compensation illustration.} Force interactions are modeled at both the end-effector and the robot base. Net forces at the end-effector are converted into positional compensation, while force compensation at the robot base is achieved through velocity adjustments.}
%     \label{fig:overview}
% \end{figure}

% Under the formulation of \cref{eq:pos_force_target}, we can easily derive several basic manipulation policies: \textit{Position Control}, \textit{Force Control}, \textit{Impedance Control} and \textit{Hybrid Position and Force Control}. In complex scenarios involving simultaneous position commands, force commands, and external disturbances, the system adheres to \cref{eq:pos_force_target}, integrating these basic control modes. A detailed explanation and formulation of these basic control modes are provided in~\cref{sec:appendix:formulation} for completeness.

\paragraph{Multi-contact Modeling}\label{sec:method:base_formulation} 
For other robot body parts beyond the end-effector, the formulation in~\cref{eq:pos_force_target} can be extended accordingly. Taking the robot base as an example, we typically care not about its joint state, but rather its velocity or global position. In such cases,~\cref{eq:pos_force_target} can be simplified by assuming the robot is controlled through base velocity commands. Specifically, given the velocity and force commands $\rvv_{\text{base}}^{\text{cmd}} = \dot{\rvx}_{\text{base}}^{\text{cmd}}$ and $\mF^{\text{cmd}}_{\text{base}}$, along with external disturbances $\mF^{\text{ext}}_{\text{base}}$, we can derive from \cref{eq:impedance_general}:
    \begin{equation}
        \mF_{\text{base}} = D(\dot{\rvx}_{\text{base}} - \dot{\rvx}_{\text{base}}^{\text{des}}) = D(\rvv_{\text{base}} - \rvv_{\text{base}}^{\text{des}}),
        \label{eq:base_simplification}
    \end{equation}
where we omit the position term because global position of the base is not available.
Here, $\mF_{\text{base}} = \mF_{\text{base}}^{\text{ext}} + (\mF_{\text{base}}^{\text{cmd}} - \mF_{\text{base}}^{\text{react}})$ is the net force and transform \cref{eq:pos_force_target} into:
\begin{equation}
    \begin{aligned}
    % \dot{\rvx}_{\text{base}}^{\text{target}} & = \dot{\rvx}^{\text{cmd}}_{\text{base}} + \frac{F_{\text{base}}^{\text{ext}} + F_{\text{base}}^{\text{cmd}}}{D}, \\
    \rvv_{\text{base}}^{\text{target}} & = \rvv^{\text{cmd}}_{\text{base}} + \frac{\mF_{\text{base}}^{\text{ext}} + (\mF_{\text{base}}^{\text{cmd}} - \mF_{\text{base}}^{\text{react}})}{D}.
    \end{aligned}
    \label{eq:base_pos_force}
\end{equation}
After this transformation, we can implement similar basic control modes using derivations from \cref{eq:base_pos_force}. Furthermore, this formulation can be extended to scenarios where external disturbances and force commands on body parts are transformed to end-effectors by converting the net force on the robot base $\mF_{\text{base}}$ into an external force to the end-effector $\mF_{\text{base}2{\text{ee}}}$, and \textit{vice versa}. However, due to the learning complexity of such methods, this work focuses on treating the end-effector and robot base independently, leaving the integrated derivation as important future work.

 As a summary, we build our policy learning with reward provided following \cref{eq:pos_force_target,eq:base_pos_force} which models the behavior of the end-effector and the base of the legged robot considering both active and passive forces. We provide the detailed training settings and model in~\cref{sec:learning}.
 
%===============================================================================

\subsection{Learning a Unified Force-Position Control Policy} \label{sec:learning}

We detail the learning of the proposed unified force-position control policy by first defining the space of observations, commands, and actions.
% In this section, we detail the learning of the proposed unified force-position whole-body control policy. This policy generates joint position targets for both the legs and arms based on input force and position commands and considering disturbance forces following derivations from \cref{eq:pos_force_target,eq:base_pos_force} in~\cref{sec:method:formulation}.
% \paragraph{Observation, Command, and Action}
% \label{sec:learning:obs_act}
Specifically, we define the robot's observation $\rvo_t$ with the robot's base orientation $\rvg^{\text{base}}_t$, angular velocity $\omega^{\text{base}}_t$, joint position $\rvq_t$, joint velocities $\dot{\rvq}_t$, previous action $\rva_{t-1}$, command $\rvc^{\text{cmd}}_t$, and the feet clock timings $\theta^{\text{feet}}_t$: 
\begin{equation}
\rvo_t = \left[\rvg^{\text{base}}_t, \rvomega^{\text{base}}_t, \rvq_t, \dot{\rvq}_t, \rva_{t-1}, \rvc^{\text{cmd}}_t, \rvtheta^{\text{feet}}_t \right]
\end{equation}
where the input command $\rvc^{\text{cmd}}_t = [\rvv_{\text{base}}^{\text{cmd}}, \rvx_{\text{ee}}^{\text{cmd}}, \mF_{\text{ee}}^{\text{cmd}}, \mF_{\text{base}}^{\text{cmd}}]$ covers the base velocity, end-effector position, end-effector force, and base force commands. For quadrupedal robots, we consider all four command types. For humanoid robots, only the locomotion command $\rvv^{\text{cmd}}_{\text{base}}$ and base force command $\mF^{\text{cmd}}_{\text{base}}$ are considered as there is no gripper available on the robot for manipulation tasks. The output action $\rva_t$ is a residual added to a predefined default pose and $\rvq^{\text{target}}_t$ are the joint position targets for the PD controller, calculated as 
$\rvq^{\text{target}}_t = \sigma_a \rva_t + \rvq^{\text{default}}$,
where $\sigma_a$ scales the policy output and $\rvq^{\text{default}}$ represents a standard pose. 

\paragraph{Policy Design} \label{sec:learning:policy_design}
We provide an overview of our policy model in~\cref{fig:method:model}(a). Our policy model comprises three modules: the observation encoder, the state estimator, and the actor. The encoder processes the observation history $\rvo_{[t-H, \dots, t-1, t]}$ $(H = 32)$ and outputs a latent feature, which is then sent to the state estimator and the actor. The state estimator then predicts the robot's state, including the external force $\mF = \mF^{\text{ext}} + \mF^{\text{react}}$, the end-effector position, and the base velocity. This estimated force could then be translated to command signals in certain desired control behaviors. 
% We utilize \ac{ppo}~\cite{schulman2017proximal} to train the actor policy and implement the state estimator with a \ac{mlp} network for state prediction outputs.

% \bx{Using the proposed formula \cref{eq:pos_force_target}, $\mF_{cmd}$ can be converted into a positional offset and compensated at the end-effector's actual position when no external force is present. This approach prevents the end-effector from continuously moving unnecessarily. Furthermore, in scenarios with obstacles, the external force equals the reaction force of the force command. In this case, the end-effector remains at the desired position, stably located on the surface of the object, while applying the required force, thereby achieving the expected behavior.}

\vspace{-5pt}

\paragraph{Force Simulation}
\label{sec:learning:policy_learning}
To simulate diverse scenarios for learning the unified force-position policy, we randomly sample position, velocity, and force commands, along with external net forces as required in~\cref{eq:pos_force_target,eq:base_pos_force}. The ranges of input commands and external forces are detailed in~\cref{sec:appendix:cmd_details}. Notably, the reaction force $\mF^{\text{react}}$ is not modeled explicitly but is incorporated into the net external force $\mF = \mF^{\text{ext}} + \mF^{\text{react}}$. 
The sampling range of $\rvx^{\text{cmd}}$ slightly exceeds the arm's original workspace without whole-body movement, while ensuring that the resulting $\rvx^{\text{target}}_{\text{ee}}$ remains within the operational limits when whole-body motion is allowed.
During training, as illustrated in~\cref{fig:method:model}(d), sampled forces are linearly ramped up to target values, held constant for a fixed interval, and then reduced back to zero according to a pre-defined schedule. After a brief zero-force period, new forces are and the cycle repeats. This sampling strategy exposes the policy to a variety of control conditions, echoing the different desired control behavior discussed in~\cref{sec:method:formulation} and enabling a single policy to adapt to varying control task demands. 

\paragraph{Policy Learning} We adopt a two-stage training procedure: first focusing on whole-body reaching and locomotion, then introducing random force commands and external disturbances. This staged approach empirically yields more stable training than a single-stage setup, as further analyzed in~\cref{sec:appendix:rl_training_details}. Policy learning is supervised by rewarding accurate tracking of the target end-effector position $\rvx_{\text{ee}}^{\text{target}}$ and base velocity $\rvv_{\text{base}}^{\text{target}}$ under varying input and disturbance combinations. Additionally, an MSE loss is used to improve the accuracy of the state estimator for both robot state and external force. Full reward specifications are provided in~\cref{tab:method:rewards}.

% \paragraph{Learning Objectives}
% Following \cref{eq:pos_force_target,eq:base_pos_force}, we supervise the learning process by rewarding the achievement of the expected target position of the end-effector $\rvx_{\text{ee}}^{\text{target}}$ and the target velocity of the robot base $\rvv_{\text{base}}^{\text{target}}$ under different combinations of input commands and external forces. In addition, we supervise the state estimator with an MSE loss to enhance its accuracy in both robot state and external force estimation. Due to the page limit, we provide all reward functions used in~\cref{tab:method:rewards}.

%===============================================================================

\subsection{Force-aware Imitation Learning}\label{sec:learning:imitation}

Recognizing the importance of force information in real-world manipulation tasks and its absence in most existing datasets, we leverage our learned force-position policy to collect force-aware data for imitation learning. Concretely, we teleoperate the robot to record joint states, base states, control commands, estimated end-effector contact forces, and RGB images from cameras mounted on both the end-effector and the robot base. This data is used to train a diffusion-based force-aware imitation learning policy that takes as input the robot states, estimated forces, and image observations, and predicts both force and end-effector position commands as inputs to our low-level force-position policy. Unlike prior works relying solely on visual inputs, our force estimator supplements the policy with contact information, enabling more accurate object interaction and force application. We validate the impact of the collected data and the effectiveness of our approach in~\cref{sec:exp:imitation}. Details of the teleoperation pipeline and training procedures are provided in \cref{sec:appendix:teleop_details} and \cref{sec:appendix:il_training_details}.
\vspace{-0.5em}

\section{Experiment}

\subsection{Force and Position Command Tracking}

\begin{figure*}[t!]
    \centering
    \includegraphics[width = \linewidth]{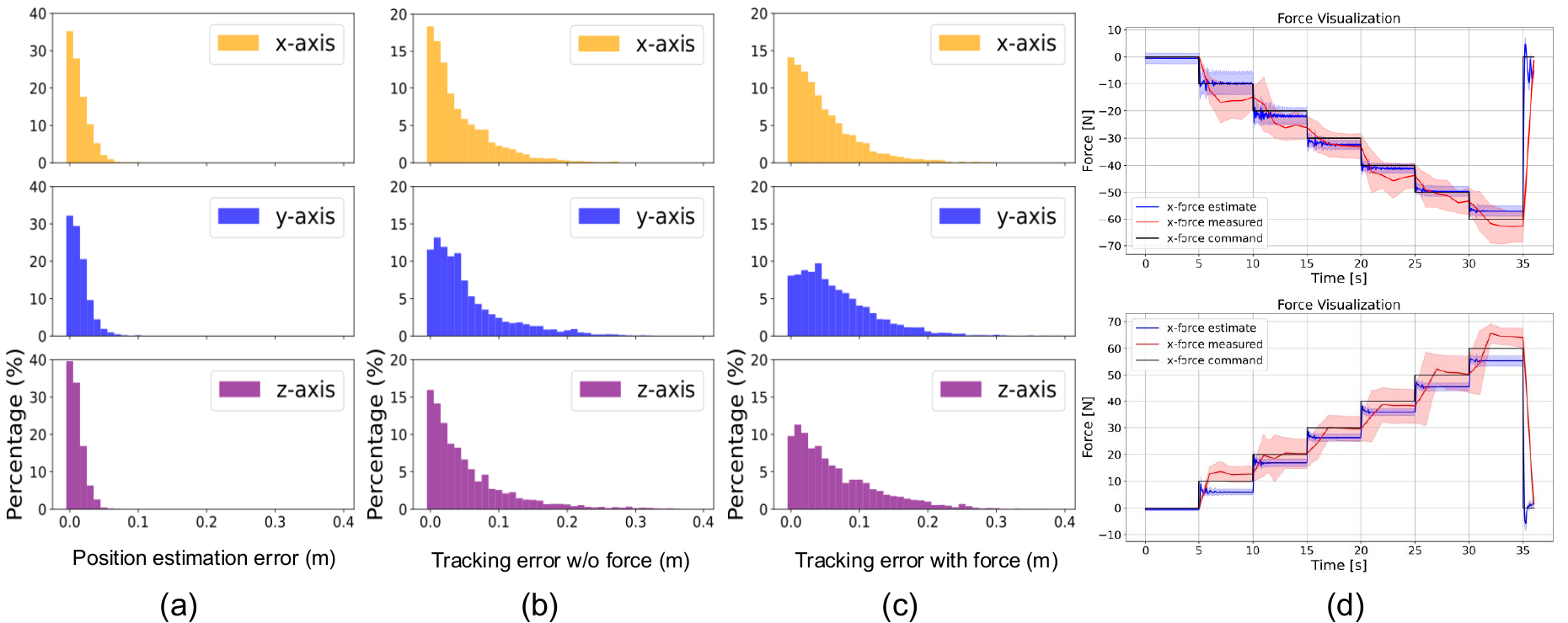}
    \vspace{-15pt}
    \caption{\textbf{Force and position control evaluation.} (a)–(c) Evaluation of force and position control tracking errors in simulation environments. (d) Real-world evaluation of force control, shaded areas indicate variance measured across 5 different end-effector positions.}
    \label{fig:unified_experiment}
    \vspace{-10pt}
\end{figure*}

\paragraph{Position Tracking}
To evaluate performance in simulation, we conduct 6000-step rollouts with randomly generated end-effector trajectories with position commands only, covering the entire training workspace. We report the average position tracking and estimation errors across these trials. As shown in~\cref{fig:unified_experiment}(b), the end-effector tracking error remains mostly within 0.1m when in the absence of external forces and force commands. Slightly higher errors are observed along the Y-axis, likely due to fewer available degrees of freedom in that direction, which limits precision. We also assess the accuracy of the state estimator by comparing the estimated end-effector positions with ground-truth simulation values. Across all axes, the estimation error remains within 0.05m, as shown in~\cref{fig:unified_experiment}(a).

\paragraph{Force Control}
We evaluate the ability of our proposed policy to estimate and act with forces under two settings. First, we assess position tracking performance when the policy receives force commands that match the applied external forces, serving as an indirect evaluation for evaluating unified force-position control. As shown in~\cref{fig:unified_experiment}(c), compared to the experiment without external forces, tracking error increases slightly compared to the no-force setting but remains mostly within 0.1m, demonstrating effective force-aware behavior. Second, we conduct a direct force control evaluation on real robots by applying force commands ranging from 0 N to 60 N and measuring end-effector forces using a dynamometer. Measurements at five different end-effector positions yield average errors within 10 N, as shown in~\cref{fig:unified_experiment}(d). Force estimation across six discrete levels shows errors between 5–10 N. Due to hardware limitations, evaluations along the Y- and Z-axes are capped at 40 N. Despite minor sim-to-real discrepancies, particularly along the Y-axis, the estimator remains sufficiently accurate for the targeted manipulation tasks. More analyses are provided in~\cref{sec:appendix:real_word_force_control_experiment}.

\subsection{Force-aware Imitation Learning}\label{sec:exp:imitation}

% \begin{table}[h]
%     \tiny
%     \captionsetup{font=scriptsize}
%     \caption{Imitation learning results (\textbf{50 trials per task})}
%     % \vspace{-15pt}
%     \label{tab:rebuttal_il_performance}
%     \centering
%         \begin{tabular}{ccccc}
%         \toprule
%          Task & Wipe Blackboard & Open Cabinet & Close Cabinet & Open Drawer with Occlusion\\
%         \midrule
%         \textbf{w/o Force} & 0.22 & 0.36 & 0.30 & 0.30 \\
%         \textbf{w/ Force}  & 0.58 & 0.70 & 0.72 & 0.76 \\
%         \bottomrule
%         \end{tabular}
% \end{table}
\paragraph{Task Settings}
We evaluate our method on four real-world tasks that require hybrid force-position control and force sensing: \texttt{wipe-blackboard}, \texttt{open-cabinet}, \texttt{close-cabinet}, and \texttt{open-drawer-occlusion}. In the \texttt{wipe-blackboard} task, the robot must maintain continuous contact with the surface while moving laterally to erase ink marks. We collect $50$ trajectories and trained the force-aware diffusion policy for $30k$ steps. In \texttt{open-cabinet} and \texttt{close-cabinet}, the robot interacts with a push-to-open cabinet, while in \texttt{open-drawer-occlusion}, it opens a drawer that gradually becomes occluded in visual observations (see the setup in~\cref{fig:open_drawer}). For each of the three open/close tasks, we collect $30$ episodes per task and train the policy for $20k$ steps.
As a baseline, we deploy the trained low-level policy but exclude the force estimator and force command signals during teleoperation data collection. Each task is performed 50 times with each trial constrained to a maximum of $1000$ steps ($\sim$20\ seconds) for successful completion. For completeness, we provide additional details about the task definitions and experimental settings in~\cref{sec:appendix:il_training_details}.

\begin{figure*}[t!]
    \centering
    \includegraphics[width=0.95\linewidth]{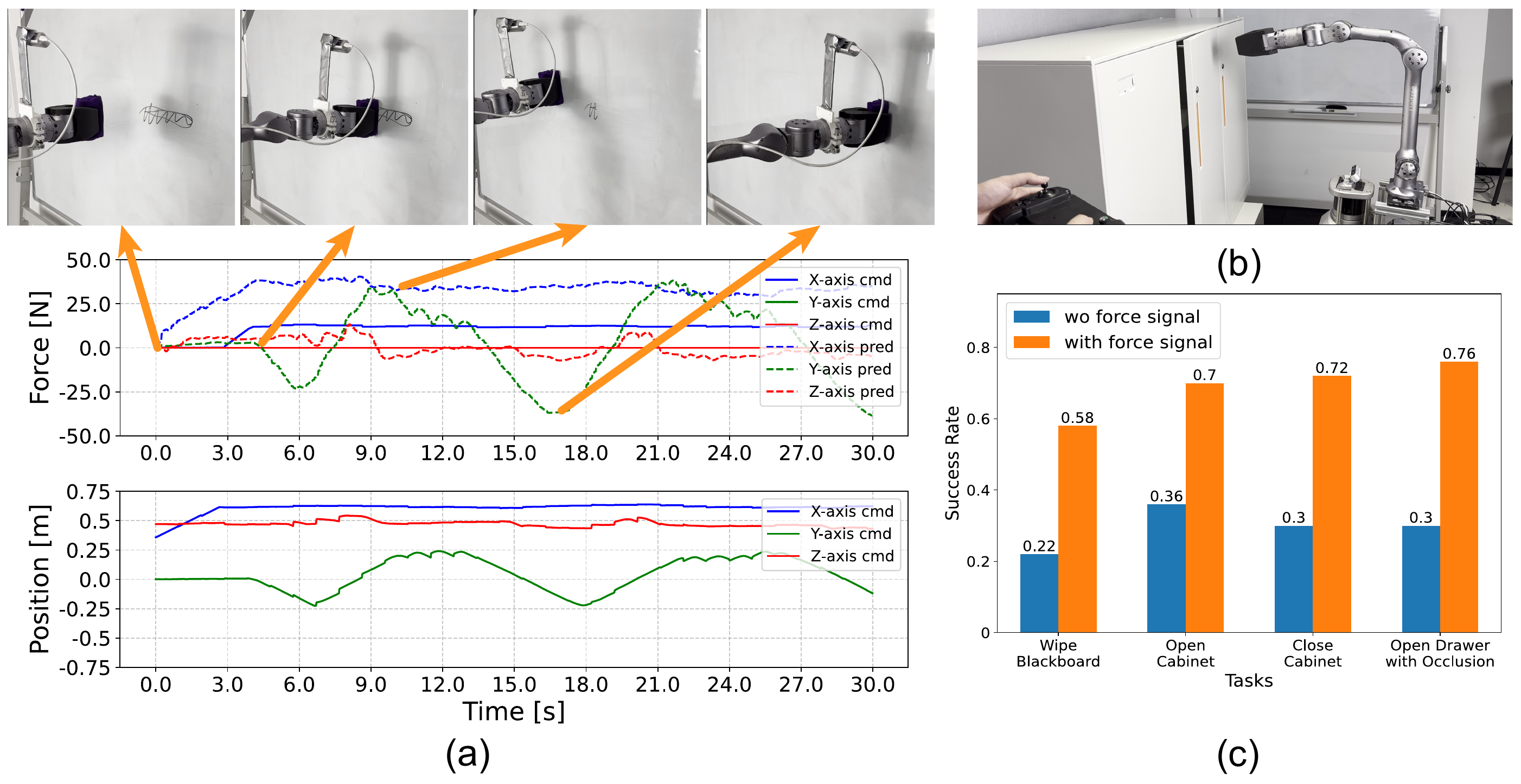}
    \vspace{-5pt}
    \caption{\textbf{Force-aware imitation learning.} (a) Time-series outputs of position and force commands to the trained force-aware imitation policy in the \texttt{wipe-blackboard} task. \textit{cmd} denotes the output of the imitation learning policy, while \textit{pred} indicates the external force estimated by the low-level policy. (b) A visualization of the data collection process. (c) The performance comparison between our policy and a baseline vision-only policy over 50 trials across four tasks.}
    \label{fig:imitation}
    \vspace{-15pt}
\end{figure*}
\paragraph{Results and Analyses}
In~\cref{fig:imitation} , we compare our method to the baseline in the four real-world tasks. Our approach achieves $\sim$39.5\% higher success rates than the baseline.
In \texttt{wipe-blackboard}, the position-only policy fails to maintain stable contact, often resulting in insufficient wiping or excessive force that risks surface damage. In contrast, our force-aware policy ensures consistent contact pressure, while the low-level policy improves compliance and reduces mechanical stress.
For \texttt{open-} and \texttt{close-cabinet}, the primary challenge lies in the push-to-open mechanism's narrow 3mm stroke, which is difficult to detect using vision alone. Our force estimator accurately senses the required contact force, enabling reliable activation. In \texttt{open-drawer-occlusion}, the baseline policy, relying solely on visual cues, suffers a sharp success rate drop to 0.3 due to unobservable contact. Our method leverages force sensing to detect contact under occlusion, increasing the success rate to 0.76 and underscoring the importance of force estimation in vision-compromised scenarios. We provide all quantitative comparisons and additional details in \cref{sec:appendix:il_training_details}.

\subsection{Basic Manipulation Policies}
\begin{figure*}[t!]
    \centering
    \includegraphics[width=\linewidth]{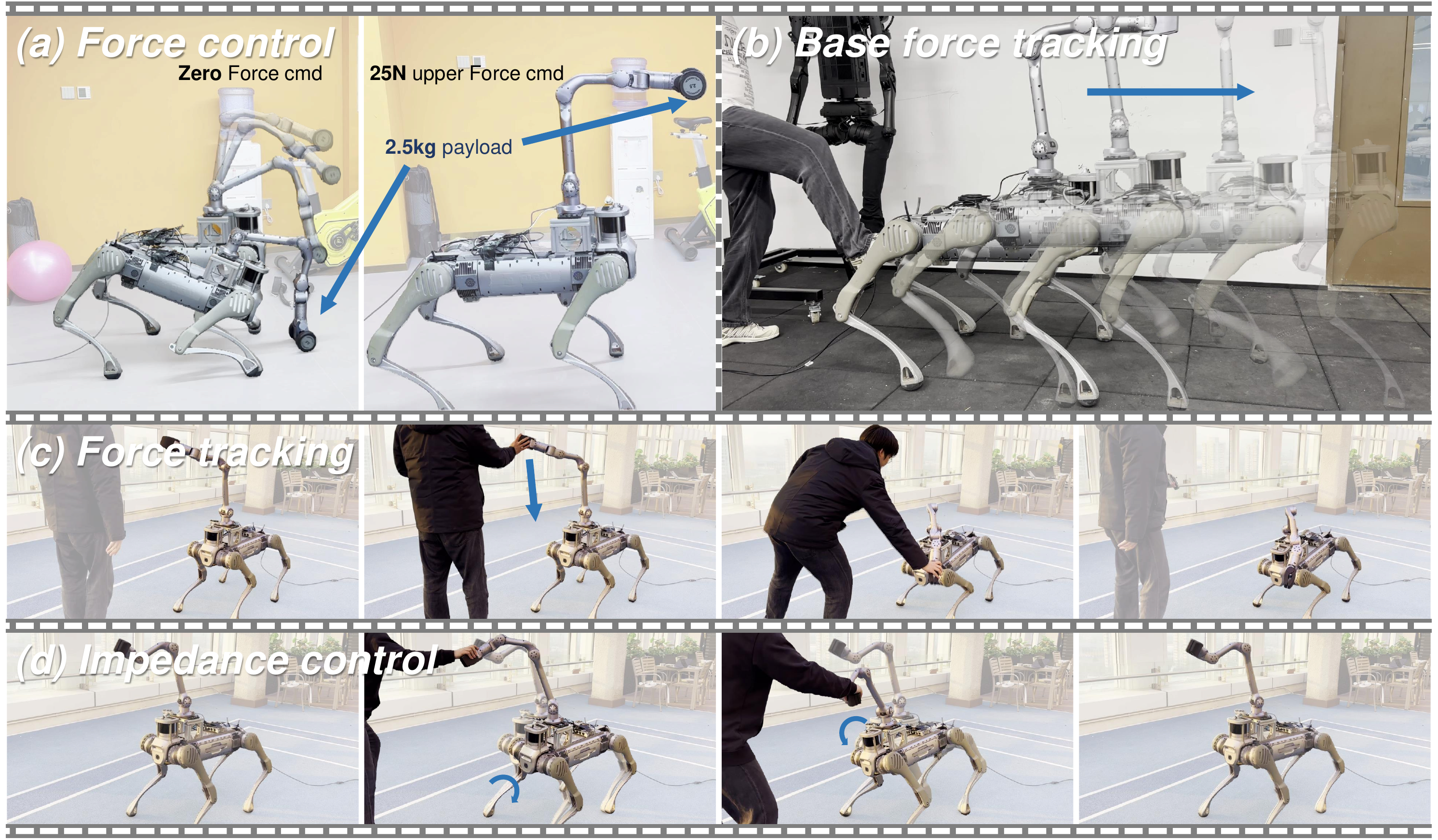}
    \caption{\textbf{Diverse skills facilitated by our policy.} (a) Force control: The robot counteracts gravity to support a payload when given a $25N$ force command. (d) Base force tracking: The robot responds compliantly to pushes on its base, enabling intuitive human guidance. (c) Force tracking: The robot tracks a zero-force command by minimizing external force interactions. (d) Impedance control: The robot adjusts its whole-body posture to counteract and comply with external disturbances.}
    \label{fig:skills}
    \vspace{-15pt}
\end{figure*}

\paragraph{Force Control}
Force control directly applies a commanded force by moving the end-effector in the force direction until the applied and commanded forces match. Our unified strategy requires no additional training; following ~\cref{eq:force_control}, where the sum of the estimated external force and the force command determines displacement compensation until equilibrium is achieved. As demonstrated in ~\cref{fig:skills}(a) and the supplementary video, when a $2.5 kg$ dumbbell is attached to the end-effector, the robot achieves balance with a $25 N$ upward force command. Without this command, however, the end-effector drops due to the dumbbell's gravity.
\vspace{-0.10in}

\paragraph{Force tracking}
Force tracking is a special case of force control where the end-effector tracks a zero-force command. When external forces are applied, the end-effector moves in the force direction. Once the force is removed, it stays in the displaced position instead of returning to the original target. This behavior is implemented using our unified policy, following ~\cref{eq:force_tracking}. As shown in ~\cref{fig:skills}(c) and the supplementary video, this capability is demonstrated by setting the force command to zero. In this case, the end-effector follows the external force and remains in the displaced position after the force is removed, achieving effective force tracking.
\vspace{-0.10in}

\paragraph{Impedance Control}
A key application of force control is impedance control, where the end-effector tracks a target position while responding compliantly to external forces, following the dynamics of a spring-mass-damper system. We implement impedance control using the unified policy, following ~\cref{eq:impedance_control}. As shown in ~\cref{fig:skills}(d) and the supplementary video, we demonstrate this capability in human-robot tug-of-war and arm-wrestling scenarios. In these tasks, the further the end-effector deviates from the target position, the greater the resistive force exerted by the robot, showcasing impedance behavior.
\vspace{-0.05in}

\subsection{Cross Embodiment Performance}
\vspace{-0.05in}
To validate the cross-embodiment capability of our unified policy,we tested it on the Unitree G1 humanoid robot and Unitree B2-Z1 quadrupedal manipulator. For locomotion, unlike manipulation tasks where force compensation applies directly to the end-effector, we adjust the robot base velocity to compensate for external forces, following ~\cref{eq:base_pos_force}.
As shown in the third row of ~\cref{fig:teaser}, when the compensated velocity equals and opposes the velocity command, the humanoid robot halts and leans its body to maintain balance. Similarly, as shown in ~\cref{fig:skills}(b), the quadrupedal robot begins walking forward when kicked, even with zero force and velocity commands.
\vspace{-0.05in}

%===============================================================================

\section{Conclusion}
\vspace{-0.05in}
We propose a unified force-position control policy for legged robots, enabling contact-rich loco-manipulation tasks without explicit force sensors. Using reinforcement learning, our policy estimates external forces from historical states and compensates for them through position and velocity adjustments. This approach supports diverse behaviors like position tracking, force application, and compliance. Additionally, integrating force estimation into imitation learning improves task success in contact-rich environments. Experiments on quadrupedal and humanoid robots validate the policy’s adaptability and robustness in real-world scenarios.

\clearpage

\section{Limitations and Future Work}

% Despite its promising performance, our approach has certain limitations.

First, while the policy successfully estimates external forces without direct force sensing, its accuracy tends to degrade in high-frequency interactions and at the edges of the robot’s workspace. Future work could focus on improving force estimation in these corner cases. One possible direction is to incorporate velocity and acceleration terms from \cref{eq:pos_force_target} to enhance force estimation, allowing the model to better capture dynamic interactions.

Second, while our policy generalizes well from simulation to real-world deployment, discrepancies remain due to the sim-to-real gap, particularly in force accuracy along different coordinate axes. These differences likely stem from mismatches in actuator dynamics and contact modeling between simulation and real hardware. Future work could explore techniques such as domain randomization and real-to-sim corrections to improve robustness across varying real-world conditions.

Additionally, our current framework primarily focuses on estimating force at a single interaction point. Future work could explore multi-point force estimation and whole-body force interaction tasks. For example, in scenarios such as a quadrupedal robot opening a heavy door, the robot could use its body to brace against the door while simultaneously using its manipulator to press down on the handle. Developing policies that coordinate multiple contact forces across different body parts could enable more complex and effective real-world interactions.

%===============================================================================

%===============================================================================

% \clearpage
% \bibliographystyle{plainnat}
\bibliography{references}

%===============================================================================

% The acknowledgments are automatically included only in the final and preprint versions of the paper.
% \acknowledgments{If a paper is accepted, the final camera-ready version will (and probably should) include acknowledgments. All acknowledgments go at the end of the paper, including thanks to reviewers who gave useful comments, to colleagues who contributed to the ideas, and to funding agencies and corporate sponsors that provided financial support.}

%===============================================================================

% \bibliography{example}  % .bib

%appendix
%实验场景的图，training detail，reward curve，reward，
\clearpage
\appendix \label{sec:appendix}

\renewcommand{\thefigure}{A.\arabic{figure}}
\renewcommand{\thetable}{A.\arabic{table}}
\renewcommand{\theequation}{A.\arabic{equation}}

\section{Problem Formulation} \label{sec:appendix:formulation}
Our policy enables a wide range of manipulation behaviors under varying force and position inputs, including position control, force control, impedance control, and hybrid position and force control. Under the formulation of \cref{eq:pos_force_target}, we can easily derive the following basic manipulation policies:
\begin{itemize}
    \item[\textit{a)}] {
        \textit{Position Control}: When there is no external disturbance or active force command applied, \cref{eq:pos_force_target} becomes 
        \begin{equation}
            \rvx^{\text{target}} = \rvx^{\text{cmd}},
            \label{eq:pos_control}
        \end{equation}
        where the target end-effector position $\rvx^{\text{target}}$ should reach the commanded position $\rvx^{\text{cmd}}$.
    }
    \item[\textit{b)}] {
        \textit{Force Control}: When in contact with the environment and applying a force $\mF^{\text{cmd}}$ without external disturbances, the desired goal position of the end-effector is defined following \cref{eq:pos_force_target} as:
        \begin{equation}
            \begin{aligned}
                \rvx^{\text{target}} = \rvx^{\text{cmd}} + \frac{(\mF^{\text{cmd}} - \mF^{\text{react}})}{K}.
            \end{aligned}
            \label{eq:force_control}
        \end{equation}
        During the system execution, the reaction force $\mF^{\text{react}}$ gradually increases to match the force command $\mF^{\text{cmd}}$, leaving the final target pose $\rvx^{\text{target}}_\text{final} = \rvx^{\text{cmd}}$.
    }
    \item[\textit{c)}] {
        \textit{Impedance Control}: When the end-effector is subjected to an external disturbance force but does not apply force to the environment, \cref{eq:pos_force_target} simplify to:
        \begin{equation}
            \rvx^{\text{target}} = \rvx^{\text{cmd}} + \frac{\mF^{\text{ext}}}{K},
            \label{eq:impedance_control}
        \end{equation}
        where the end-effector, when subjected to external disturbances, adjusts its position to exhibit compliance in response to the external force $\mF^{\text{ext}}$. Notably, we can also implement force tracking and gravity compensation using \cref{eq:impedance_control} by dynamically adjusting the position command $\rvx^{\text{cmd}}$ following:
        \begin{equation}
            \Delta \rvx^{\text{cmd}} = \frac{\mF^{\text{ext}}}{K},
            \label{eq:force_tracking}
        \end{equation}
        where $\mF^{\text{ext}}$ can be the gravity term or an external force.
        }
    \item[\textit{d)}] {
        \textit{Hybrid Position and Force Control}: As defined in~\cite{raibert1981hybrid}, hybrid position and force control refers to controlling the end-effector's movement using $\rvx^{\text{cmd}}$ while applying a force command $\mF^{\text{cmd}} = \mF^{\text{cmd}}_{\perp}$ perpendicular to the tangential direction of the movement. In this scenario, the system follows \cref{eq:pos_control} along the tangential direction without force command and satisfies \cref{eq:force_control} in the perpendicular direction where the force command is active.
    }
\end{itemize}

\paragraph{Simplified impedance model.}In our formulation, the damping and inertia terms are omitted. This simplification is suitable for relatively static or quasi-static tasks (e.g., wiping or tug-of-war), where motion is slow and normal force ensures contact. In these cases, the static model is sufficient. For dynamic or agile skills, higher control frequencies and explicit velocity/acceleration terms will be necessary, which we plan to explore in future work.

\paragraph{Assumption of rigid contact.}Our formulation \(\mF = K(\rvx - \rvx^{\text{des}})\) naturally handles both rigid and compliant objects, as the end effector position depends only on the net force $\mF$ and the desired goal position $\rvx^{\text{des}}$, regardless of the stiffness of the object. For rigid contacts, small displacements generate the required force, whereas for soft objects, larger deformations occur until the force equilibrium is reached.

\section{Hardware Settings and Teleoperation System} \label{sec:appendix:teleop_details}

\begin{figure*}[t!]
    \centering
    \begin{subfigure}[b]{0.48\linewidth}
        \includegraphics[width=\linewidth]{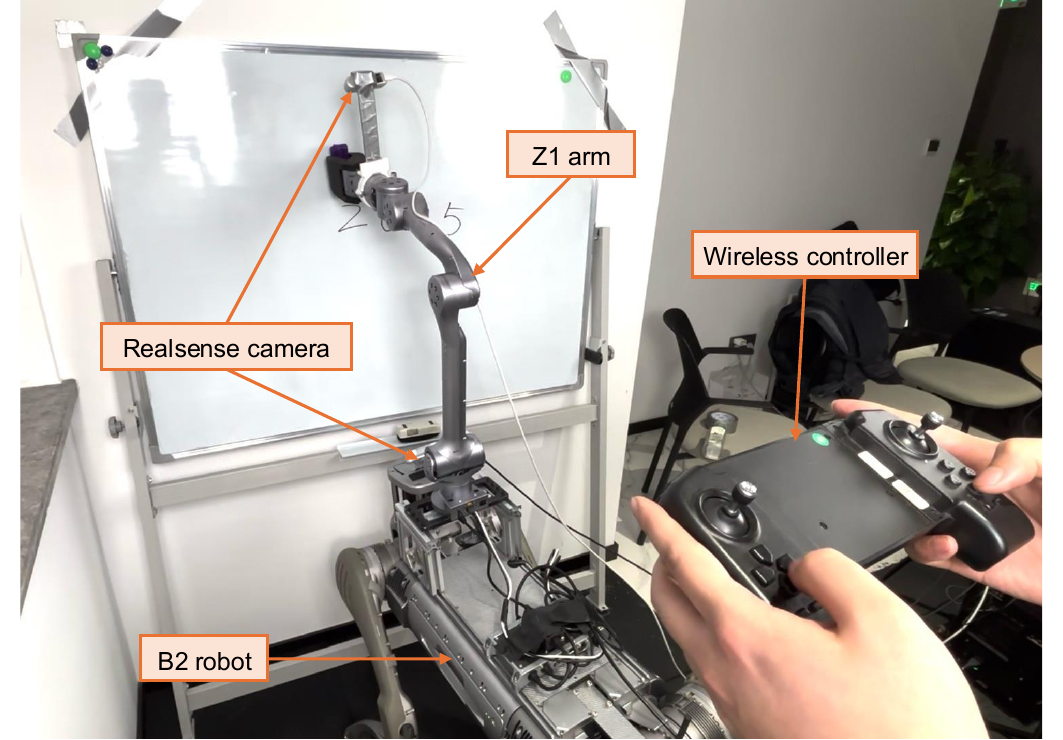}
        \caption{\textbf{B2-Z1 robot hardware.} A Unitree B2 robot with a Z1 arm is teleoperated via a wireless controller, with two RealSense cameras for visual input.}
        \label{fig:b2z1_hardware}
    \end{subfigure}
    \hspace{0.0\linewidth}
    \begin{subfigure}[b]{0.48\linewidth}
        \includegraphics[width=\linewidth]{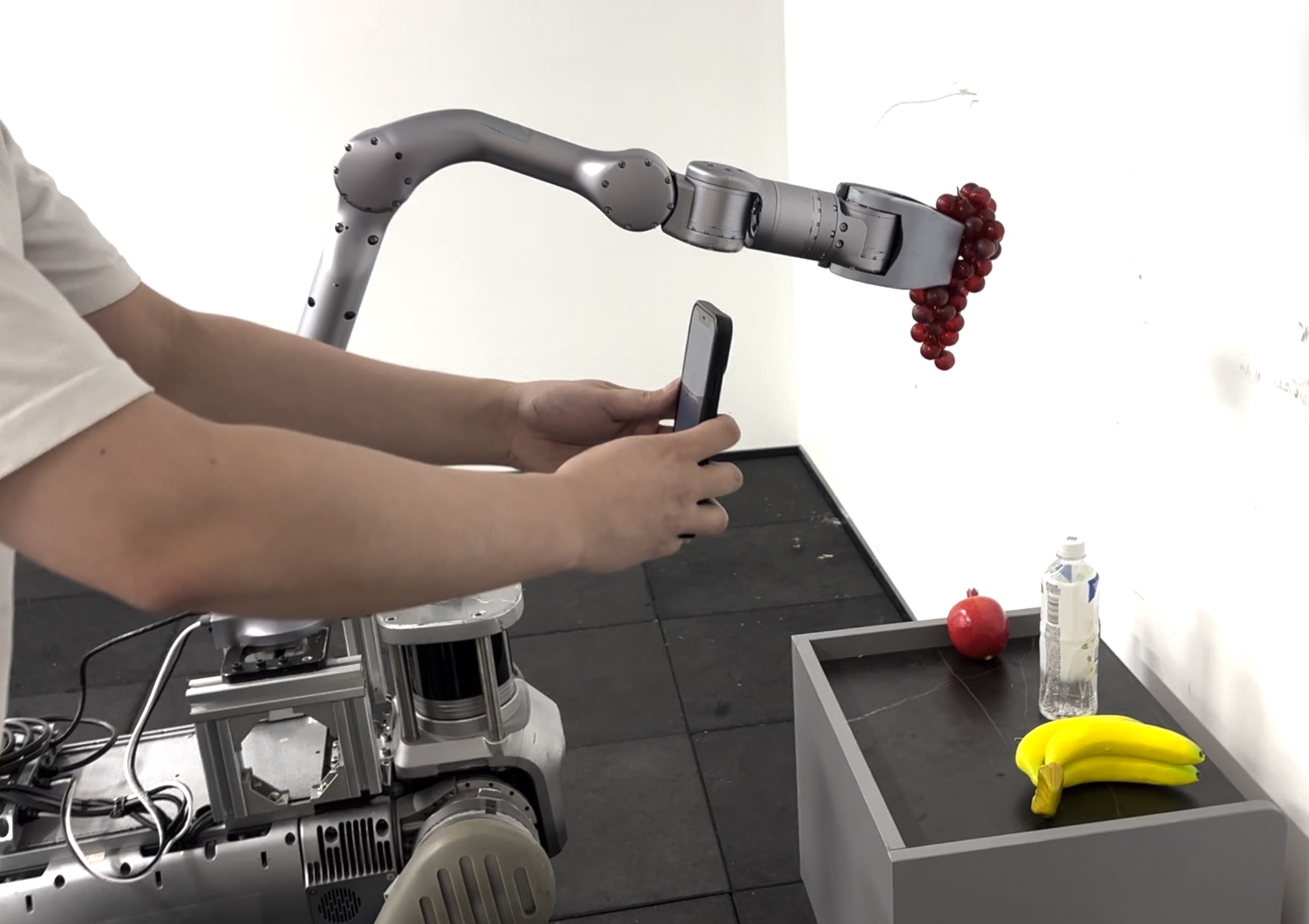}
        \caption{\textbf{Iphone teleoperation.} Using the MuJoCo AR app, an iPhone is employed to remotely control the robot for performing manipulation tasks.}
        \label{fig:iphone_tele}
    \end{subfigure}
    \caption{\textbf{Hardware setting and teleoperation system.}}
    \label{fig:hardware}
\end{figure*}

\paragraph{Robot System Setup}
The humanoid robot system (~\cref{fig:method:model}) is the 29-DOF Unitree G1 robot. The quadruped robot system (~\cref{fig:b2z1_hardware}) consists of a 12-DOF Unitree B2 robot and a 6-DOF Unitree Z1 robot arm, both powered by the battery of B2. We customize two RealSense cameras mounted on the arm and head of the quadruped robot. Our whole-body controller and diffusion policy inference are executed on a dedicated desktop with an RTX 3090 GPU, interfaced with the B2-Z1 robot via an internet connection. 

\paragraph{Teleoperation System}
As shown in ~\cref{fig:iphone_tele}, for manipulation tasks that require only position control, we utilize the MuJoCo AR application, which allows flexible teleoperation of the robot using an iPhone. However, as illustrated in ~\cref{fig:b2z1_hardware}, this approach is no longer suitable for manipulation tasks involving hybrid force and position control. To address this, we developed a dedicated teleoperation system based on the B2 robot’s built-in wireless controller. In this system, two joysticks are used to control the robot's base movement, while eight buttons below are mapped to control the end-effector's position and the opening/closing of the gripper. Additionally, by holding the “L1” button, the system switches from issuing position commands to force commands for the end-effector.
In practice, data collection with this setup was relatively slow, e.g., approximately 20 seconds per wiping trial and 10 seconds per cabinet trial, mainly due to the limited bandwidth of teleoperation. We believe that exoskeleton-based teleoperation with force feedback will offer a more natural and efficient way to collect demonstrations for forceful manipulation, and we plan to explore this direction in future work.

\section{Details on Policy Learning with Reinforcement Learning} \label{sec:appendix:rl_training_details}

We utilize \ac{ppo}~\cite{schulman2017proximal} to train the actor policy and implement the state estimator with a \ac{mlp} network for state prediction outputs. We train our \ac{rl} policy in Isaac Gym~\cite{makoviychuk2021isaac} with $4096$ parallel environments.  

\subsection{Input Commands and Disturbance Forces} \label{sec:appendix:cmd_details}

We sample input commands and disturbance forces within the ranges below during training:
\begin{enumerate}[leftmargin=*]
\item End-effector position command in spherical coordinates within the body frame
$\rvx^{\text{cmd}}_\text{ee} = (r^{\text{cmd}}, \theta^{\text{cmd}}, \phi^{\text{cmd}})$, where 
$r^{\text{cmd}} \in [0.35, 0.85 \, \text{m}]$,
$\theta^{\text{cmd}} \in [-0.4\pi, 0.4\pi \, \text{rad}]$,
$\phi^{\text{cmd}} \in [-0.6\pi, 0.6\pi \, \text{rad}]$.
\item End-effector force command in cartesian coordinates within the body frame
\(\mF^{\text{cmd}}_{\text{ee}}\in \mathbb{R}^3\):
$[-60N, 60N]$.
\item Base velocity command $\rvv^{\text{cmd}}_{\text{base}} = (v^{\text{cmd}}_x, v^{\text{cmd}}_y, \omega^{\text{cmd}}_z)$, where 
$v_x \in [-0.8, 0.8 \, \text{m/s}]$,
$v_y \in [-0.6, 0.6 \, \text{m/s}]$,
$\omega_z \in [-0.8, 0.8 \, \text{rad/s}]$.
\item Base force command within the body frame
\(\mF^{\text{cmd}}_{\text{base}}\in \mathbb{R}^3\):
$[-60N, 60N]$.
\item External net force from the environment at the end-effector 
\(\mF_{\text{ee}}\in \mathbb{R}^3\): $[-60N, 60N]$ and at robot base \(\mF_{\text{base}}\in \mathbb{R}^3\): $[-60N, 60N]$.
\end{enumerate}

\subsection{Reward and Domain Radomization}

~\cref{tab:method:rewards} provides a detailed overview of the reward structure employed in this study, and ~\cref{tab:method:radomization} outlines the adopted domain randomization scheme.

\begin{table}[t!]
    \centering
    \caption{\textbf{Reward terms} for learning the whole-body policy.}
    \bgroup
    \def\arraystretch{1.2}
    % \scriptsize
    % \tiny
    \begin{tabular}{p{2.3cm}p{8cm}p{1.8cm}}
    \toprule
    \textit{Term} & \textit{Equation} & \textit{Weight} \\ [0.5ex]
    \hline
    \multicolumn{3}{|c|}{end-effector Unified Position and Force Control} \\
    \hline
    gripper position & $\exp\{-|\rvx_{\text{ee}}-(\rvx^{\text{cmd}} + (\mF^{\text{ext}} + \mF^{\text{cmd}} - \mF^{\text{react}})/{B}|/0.5\}$ &  {$2.0$} \\[0.5ex]

    \hline
    \multicolumn{3}{|c|}{Base Unified Position and Force Control ($\mathbf{r}_v^b$) } \\
    \hline
    base velocity  &  
     % \(\rvv_{\text{base}}^{\text{target}} & = \rvv^{\text{cmd}}_{\text{base}} + \frac{\mF_{\text{base}}^{\text{ext}} + (\mF_{\text{base}}^{\text{cmd}} - \mF_{\text{base}}^{\text{react}})}{D}.\)
    \( \exp\{-{|\rvv_{\text{base}}-(\rvv^{\text{cmd}}_{\text{base}}+F_{\text{base}}/D)| / {0.25}}\}\) & {$2.0$} \\ [0.5ex]

    \hline
    \multicolumn{3}{|c|}{Safety and Smoothness} \\
    \hline
    collision penalty & \( \mathds{1}_{\text{collision}}\) & $-5.0$ \\ [0.5ex]
    joint limit & \( \mathds{1}_{q > 0.8*q^{max} || q < 0.8*q^{min}}\) & $-10.0$ \\ [0.8ex]
    torques & \(|{\mathbf{\tau}}|^2\) & \num{-5e-6} \\ [0.5ex]
    joint velocities & \(|\dot{q}|^2\) & \num{-8e-4} \\ [0.5ex]
    joint acceleration & \(|\ddot{q}|^2\) & \num{-2e-7} \\ [0.5ex]

    action rate & \(|a_{t-1} - a_{t}|\) & \num{-0.02} \\ [0.5ex]
    torque limit &  \(\mathds{1}_{\mathbf{\tau} > 0.9*|\mathbf{\tau}^{max}|} \) & \num{-0.005}\\ [0.5ex]
     \hline
    \multicolumn{3}{|c|}{Gait}  \\
    \hline
    contact number & \( \sum_{\text{foot}}\mathds{1}_{\mathbf{\tau}_{contact}>5.} * stance\_mask \) & $2.0$ \\ [0.5ex]
    reference motion & \(|q-q^{ref}|^2\) & $1.0$ \\ [0.5ex]

    \bottomrule
    \end{tabular}
    \egroup
    \label{tab:method:rewards}
\end{table}

\begin{table}[t!]
    \centering
    \caption{\textbf{Domain radomization} for learning the whole-body policy.}
    \bgroup
    \def\arraystretch{1.2}
    % \scriptsize
    % \tiny
    \begin{tabular}{p{4cm}p{2cm}p{6cm}}
    \toprule
    \textit{Term} &  \textit{Unit}  &  \textit{Range}\\ 
    \midrule
    Friction & - &   {$[0.3,\, 2.0]$} \\[0.5ex] 
    Body Mass  & $kg$ &  {$[0.0,\, 15.0]$} \\ [0.5ex]
    base com (x,y,z axis) & m & {$[-0.15,\, 0.15]$} \\ [0.5ex]
    Motor Strength & \%  & {$[85,\, 115]$} \\ [0.5ex]
    Gripper Payload & $kg$ &{$[0.0,\, 0.5]$} \\ [0.5ex]
    Push robot & $m/s$ & {$[0.0,\, 0.8]$},  interval = 8s\\ [0.5ex]

    \bottomrule
    \end{tabular}
    \egroup
    \label{tab:method:radomization}
\end{table}

\subsection{World-Aligned End-Effector Position Estimation}

Our estimator predicts not only the external force, but also the end-effector position and base linear velocity. Although forward kinematics provides the end-effector position relative to the arm base, it decouples arm control from base posture. In contrast, we estimate the end-effector position in a world-aligned base frame (with fixed height and orientation relative to the base projection). This representation allows, for example, a downward end-effector command to naturally induce robot base leaning near workspace limits, enabling coordinated whole-body behavior without explicit base control. As the end-effector position in this frame cannot be directly measured, we estimate it instead.

\subsection{Additional Analyses on Policy Learning}
\begin{figure}[h]
  \centering
  \includegraphics[width=1.0\linewidth]{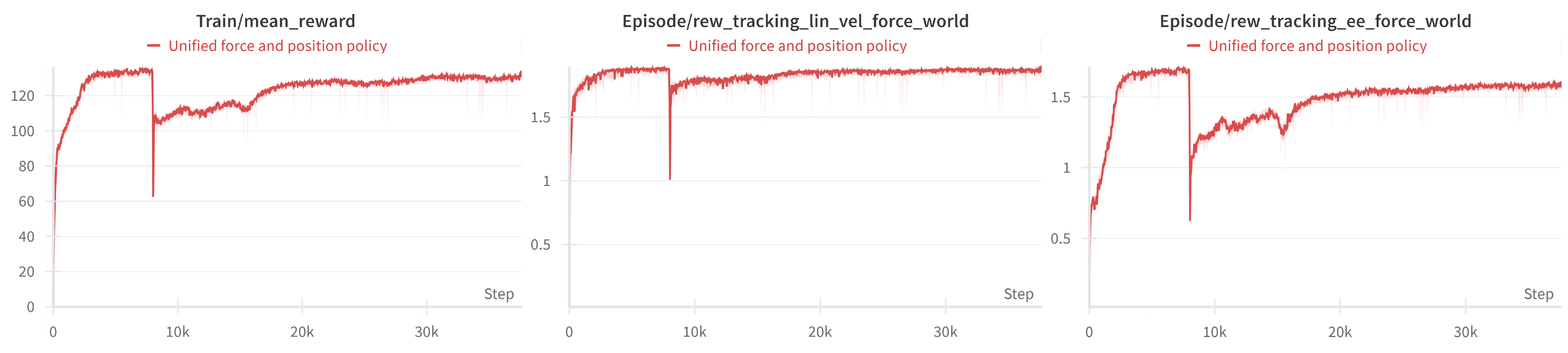}
  % \captionsetup{font=scriptsize}
  \caption{Training reward curve (view with zoom-in). \label{fig:reward_curve}}
\end{figure}
Introducing external disturbances early makes training challenging, as initial policies struggle with body balance and end-effector stability. To address this, we used a two-stage curriculum: first, training whole-body reaching and locomotion, then adding random force commands and disturbances. The training reward curve (\cref{fig:reward_curve}) shows an initial drop in stage two, with locomotion recovering and whole-body reaching stabilizing, despite a slight reaching reward decrease due to increased sampling diversity of force commands and disturbances.

\subsection{Impact of Motor Gains}

Compared to pure position-based whole-body control, we use smaller Kp/Kd gains. We observed that lower gains, which allow greater overshoot, improve force estimation. However, excessively low values degrade position tracking accuracy.

\subsection{Sim-to-real Gap and Force Estimation Robustness}

Our force estimator experiences sim2real gaps due to mismatches in motor dynamics and contact modeling. While domain randomization (e.g., varying Kp/Kd gains) helps reduce these gaps, sim2real transfer remains challenging for RL-based policies. Inspired by how humans rely more on tactile feedback than precise force sensing in contact-rich tasks, our IL policy combines estimated forces with visual input to achieve robust performance without requiring high force accuracy. To further reduce the sim2real gap, we plan to fine-tune the estimator with real-world data and apply system identification methods.

\section{Details on Force-aware Imitation Learning Policy} \label{sec:appendix:il_training_details}

\begin{table}[h]
    % \captionsetup{font=scriptsize}
    \caption{Imitation learning results (\textbf{50 trials per task})}
    % \vspace{-15pt}
    \label{tab:rebuttal_il_performance}
    \centering
    \resizebox{\linewidth}{!}{
        \begin{tabular}{ccccc}
        \toprule
         Task & \texttt{wipe-blackboard} & \texttt{open-cabinet} & \texttt{close-cabinet} & \texttt{open-drawer-occlusion}\\
        \midrule
        \textbf{w/o Force} & 0.22 & 0.36 & 0.30 & 0.30 \\
        \textbf{w/ Force}  & 0.58 & 0.70 & 0.72 & 0.76 \\
        \bottomrule
        \end{tabular}
    }
\end{table}

\begin{figure*}[t!]
    \centering
    \includegraphics[width = 0.8\linewidth]{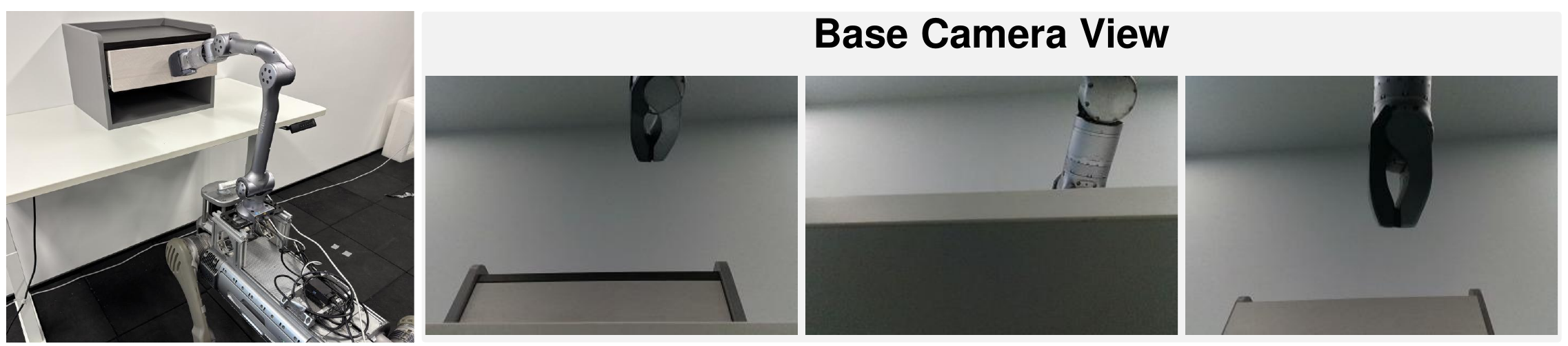}
    \caption{Open drawer with occlusion, the gripper becomes \textbf{occluded} during manipulation.}
    \label{fig:open_drawer}
\end{figure*}

\paragraph{Task Settings}
We select four tasks that require a combination of hybrid force and position control and force sensing capabilities in the real world. 
\begin{itemize}[leftmargin=*]
\item For the \texttt{wipe-blackboard} task, the robot must press against the blackboard while moving laterally to effectively remove ink marks. If the robot loses contact or fails to apply sufficient force during movement, the ink will not be wiped away. A purely position-controlled approach struggles with maintaining consistent contact force, often resulting in intermittent contact or excessive force application. In contrast, our force estimator enables the robot to sustain stable contact pressure, ensuring effective wiping while minimizing the risk of excessive force that could damage the robot. 
\item For the two tasks of \texttt{open-} and \texttt{close-cabinet} with a push-to-open cabinet, the robot must apply sufficient force to press the door, triggering the built-in mechanism that causes it to spring open or close. Unlike the previous task, this scenario requires the robot to exert enough force to overcome the mechanism’s resistance, despite the minimal displacement during activation.
\item For the task of \texttt{open-drawer-occlusion} with a rebound mechanism, the robot must apply sufficient force to press on the drawer front under visual occlusion, triggering the built-in mechanism that makes the drawer spring open, as shown in~\cref{fig:open_drawer}.
\end{itemize}

\paragraph{Evaluation} We provide the quantitative comparison between our method and the baseline in~\cref{tab:rebuttal_il_performance}. And specifically:
\begin{itemize}[leftmargin=*]
\item For the task of \texttt{wipe-blackboard}, success is defined as the robot erasing $90\%$ of the ink marks. Failure occurs if the robot exceeds the time limit without achieving this goal.

\item For the two tasks of \texttt{open-} and \texttt{close-cabinet} with a rebound mechanism, success is defined as the robot pressing the cabinet door to open (or close) the cabinet and then releasing the surface. Failure occurs if the robot is unable to activate the mechanism or does not release the cabinet door after pressing it.

\item For the task of \texttt{open-drawer-occlusion} with a rebound mechanism, success is defined as the robot pressing the drawer front under visual occlusion and then releasing the surface. Failure occurs if the robot is unable to activate the mechanism or does not release the drawer front after pressing it.
\end{itemize}

\begin{figure}[h]
    % \vspace{-10pt}
    \centering
    \begin{subfigure}[b]{0.45\linewidth}
        \centering
        \includegraphics[width=\linewidth]{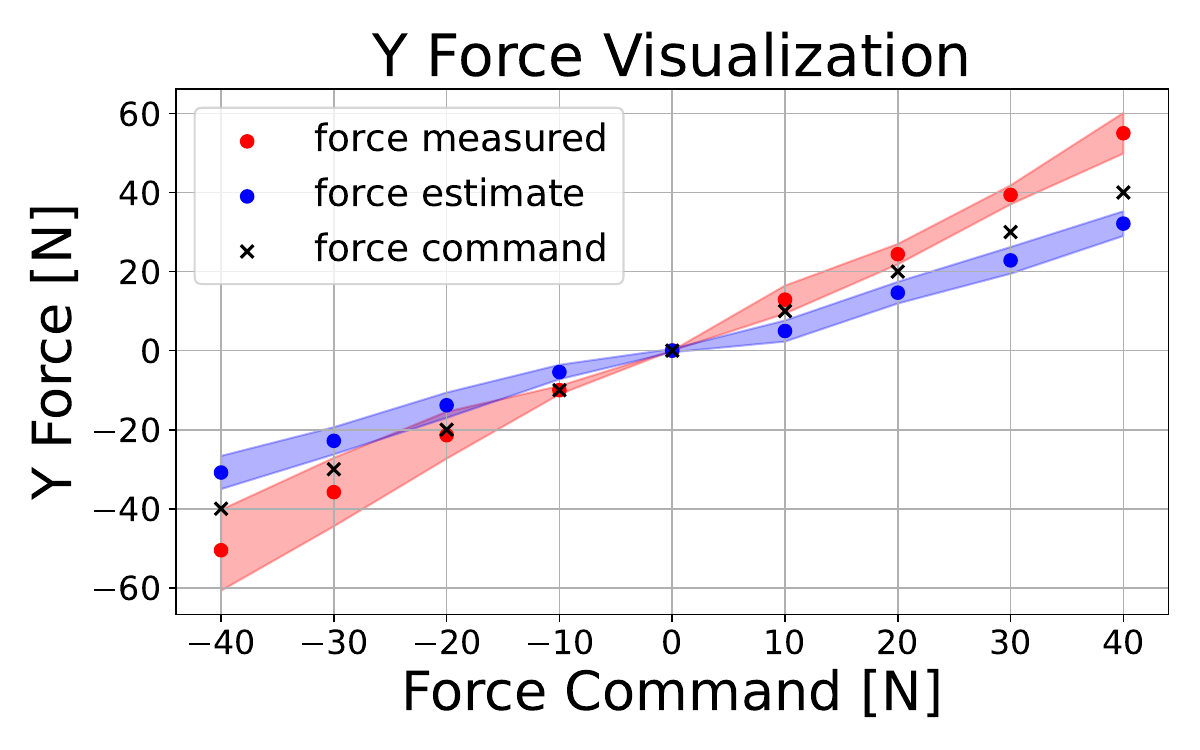}
        % \caption{$Y$ axis force error}
        \label{fig:real_pos_force: pos_est}
    \end{subfigure}
    \hspace{0.0\linewidth}
    \begin{subfigure}[b]{0.45\linewidth}
        \centering
        \includegraphics[width=\linewidth]{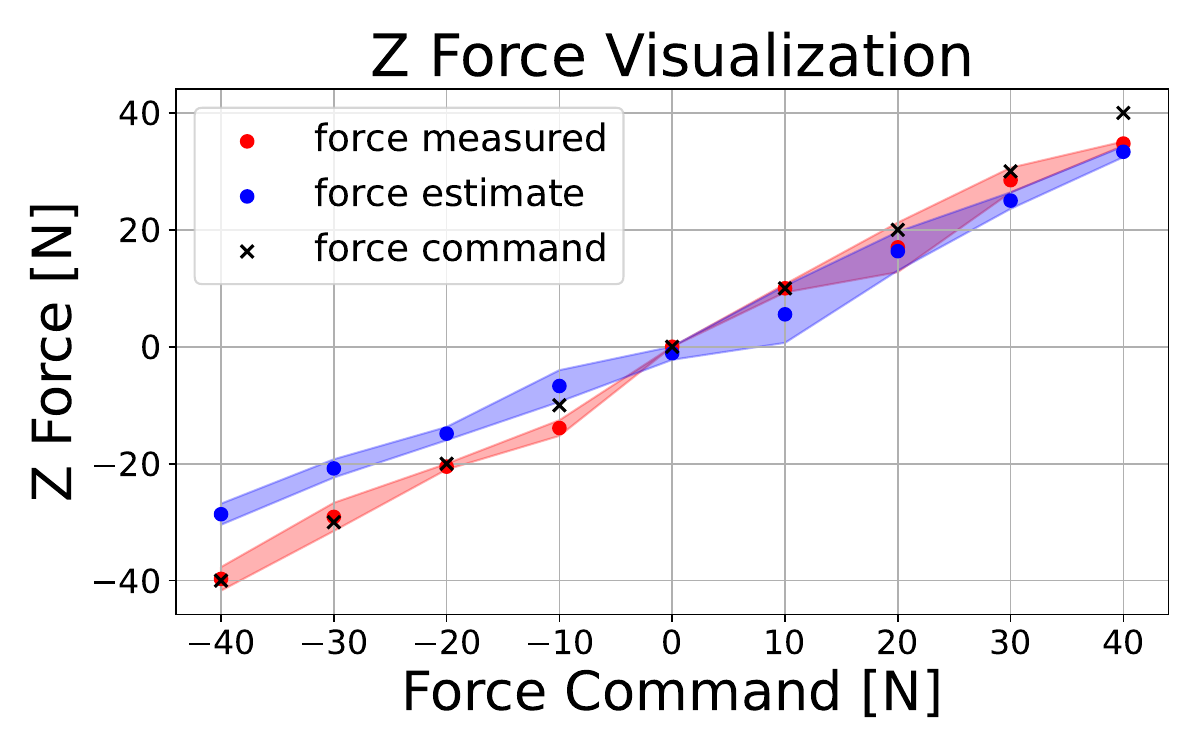}
        % \caption{$Z$ axis force error}
        \label{fig:real_pos_force: pos_tracking_woforce}
    \end{subfigure}
    % \vspace{-20pt}
    % \captionsetup{font=scriptsize}
    \caption{Real-world force control evaluation.}
    \label{fig:real_pos_force}
    % \vspace{-15pt}
\end{figure}

\section{Assessing Force Estimation Accuracy Along X and Y Axes}
\label{sec:appendix:real_word_force_control_experiment}
We evaluate the force estimator by randomly selecting five positions and applying forces ranging from -60N to 60N along the \textit{X}-axis (as in \cref{fig:imitation} of the main paper). We additionally evaluate on the \textit{Y}- and \textit{Z}-axis within 40N (due to hardware constraints of Unitree-Z1) in~\cref{fig:real_pos_force}. While sim-to-real discrepancies introduce inaccurate estimations, especially along \textit{Y}-axis, we argue that the current estimator suffices for the coarse-grained manipulation tasks discussed in this study. Reducing this sim-to-real gap for finer control will be one focus Additional of our future work.

\end{document}